\documentclass[runningheads]{llncs}
\usepackage{graphicx}
\usepackage{subfig}
\usepackage{amsmath}
\usepackage{color}
\usepackage{xcolor}
\begin{document}
%
\title{ShapeDBA: Generating Effective Time Series Prototypes using ShapeDTW Barycenter Averaging}
\titlerunning{ShapeDBA: Generating Effective Time Series Prototypes}
%
\newcommand{\hfc}[1]{{\color{red}~{\bf [Hassan:} #1{\bf ]}}}
\newcommand{\afc}[1]{{\color{blue}~{\bf [Hadi:} #1{\bf ]}}}
\newcommand{\gfc}[1]{{\color{magenta}~{\bf [Germain:} #1{\bf ]}}}
\newcommand{\mdc}[1]{{\color{teal}~{\bf [Maxime:} #1{\bf ]}}}
\newcommand{\fpc}[1]{{\color{pink}~{\bf [Francois:} #1{\bf ]}}}

\author{Ali Ismail-Fawaz\inst{1} \and
Hassan Ismail Fawaz\inst{1}\inst{4} \and
François Petitjean\inst{2} \and 
Maxime Devanne\inst{1} \and 
Jonathan Weber\inst{1} \and 
Stefano Berretti\inst{3} \and 
Geoffrey I. Webb\inst{2} \and 
Germain Forestier\inst{1,2}}
\authorrunning{A. Ismail-Fawaz et al.}
%
\institute{IRIMAS, Universite de Haute-Alsace\\
\email{first-name.last-name@uha.fr} \and
Department of Data Science and Artificial Intelligence, Monash University\\
\email{first-name.last-name@monash.edu} \and
MICC, University of Florence, Italy\\
\email{first-name.last-name@unifi.it} \and 
Ericsson Research \\\email{hassan.ismail.fawaz@ericsson.com}
}
\maketitle              
\begin{abstract}

Time series data can be found in almost every domain, ranging from the medical field to manufacturing and wireless communication.
Generating realistic and useful exemplars and prototypes is a fundamental data analysis task.
In this paper, we investigate a novel approach to generating realistic and useful exemplars and prototypes for time series data. Our approach uses a new form of time series average, the ShapeDTW Barycentric Average.
We therefore turn our attention to accurately generating time series prototypes with a novel approach.
The existing time series prototyping approaches rely on the Dynamic Time Warping (DTW) similarity measure such as DTW Barycentering Average (DBA) and SoftDBA.
These last approaches suffer from a common problem of generating out-of-distribution artifacts in their prototypes.
This is mostly caused by the DTW variant used and its incapability of detecting neighborhood similarities, instead it detects absolute similarities.
Our proposed method, ShapeDBA, uses the ShapeDTW variant of DTW, that overcomes this issue.
We chose time series clustering, a popular form of time series analysis
to evaluate the outcome of ShapeDBA compared to the other prototyping approaches.
Coupled with the $k$-means clustering algorithm, and evaluated on a total of 123 datasets from the UCR archive, our proposed averaging approach is able to achieve new state-of-the-art results in terms of Adjusted Rand Index.

\end{abstract}

\section{Introduction}
Time series data can now be seen in many real life problems.
This data is starting to be of interest in many research fields.
For instance time series can be found in medical data such as ECG signals, in human motion data, in satellite images, etc.
Generating exemplars and prototypes for time series data is an essential problem that could be used in many areas. 
For example, time series averaging is being used to generate synthetic data in order to augment the training data and boost supervised models~\cite{IsmailFawaz2018Dataaugmentationusing,Forestier2017Generatingsynthetictime} or used to make the classification task more accurate~\cite{Petitjean2014DynamicTimeWarping}.
Time series prototyping can also be used for explainability~\cite{gee2019explaining}.

One challenge when prototyping time series data is evaluation, which is addressed in most of the cases using clustering, a fundamental machine learning tool in data analysis.
Clustering is a machine learning unsupervised problem that aims to discover a set of clusters in the data that should correspond to the same distribution and the previously unseen class label.
Clustering for time series data has been very much addressed in the literature~\cite{liao2005clustering,aghabozorgi2015time}.
Varying from machine learning tools such as $k$-means and $k$-medoids~\cite{holder2022review} to the usage of deep learning~\cite{lafabregue2022end}.
Unlike other data types, basic machine learning clustering algorithms need to be adapted to the case of temporal data.
For instance, the $k$-means algorithm aims to minimize a distance between the samples in a cluster and the centroid of this cluster.
This distance is usually the Euclidean distance, but the implicit assumption using such metric is that the input samples are made of independent feature points. 
However, this is not the case in time series data, where each feature point, referred to as time stamp, is dependent with all other time stamps.
This is referred to as a temporal correlation, which obligates the definition of a replacement of the Euclidean distance in the $k$-means algorithm.
For this reason, time series similarity measures such as DTW and SoftDTW have been used instead and showed a significant improvement over the usage of the Euclidean distance. 

A further issue with the naive way of using the $k$-means algorithm, is the averaging phase to define the clusters' centroids.
The averaging method used in the $k$-means algorithm is the arithmetic mean, which presents the same problem as the Euclidean distance.
For this reason, a novel averaging method was proposed that uses the DTW similarity measure in order to produce a meaningful centroid.
This technique, DBA, showed to perform significantly better than other naive approaches.
The problem of finding a meaningful average for time series data presents much more challenges than defining the similarity metric.
This is due to the challenge in defining what an average time series does represent.
However, finding a meaningful average presents a much higher impact on the performance of the $k$-means algorithm than defining the similarity measure.
For these reasons, we address the clustering problem by producing a more respectful averaging algorithm for time series data.

The defined averaging techniques for time series data until now suffer from a common problem of generating out-of-distribution artifacts (see Figure ~\ref{fig:averages-visulization}).
This problem occurs because these averaging techniques do not look into the neighborhood of each time stamp in the time series data.
Instead, the averaging occurs after aligning each time stamp of the centroid with the ones in the time series dataset.
In this work, we propose incorporating ShapeDTW~\cite{shape-dtw} into the DBA algorithm in order to overcome this issue.
ShapeDTW is a DTW variant that manages to avoid aligning two time stamp that have closer values but in a significantly different neighborhood.
This last case study occurs often in time series data and is the main reason, to the best of our knowledge, for the existence of the generated artifacts.
The ShapeDTW similarity measure coupled with DBA, i.e., the proposed ShapeDBA algorithm, is coupled with the $k$-means algorithm in order to apply clustering on time series data.

The contributions of this work are:
\begin{itemize}
    \item Proposing a novel averaging algorithm ShapeDBA based on ShapeDTW;
    \item Extensive experiments on the UCR archive showing that ShapeDBA achieves state-of-the-art performance following the Adjusted Rand Index metric;
    \item Efficient implementation of ShapeDTW resulting in ShapeDBA being faster than SoftDBA.
\end{itemize}


\section{Related Work}

\subsubsection{Definitions}
The following definitions will be used throughout the rest of the paper:
\begin{itemize}
    \item Univariate Time Series (UTS) $\textbf{x}=\{x_0,x_1,\ldots, x_{L-1}\}$ is a sequence of length $L$ made of correlated data points equally separated in time.
    \item A TSC dataset $\mathcal{D}=\{(\textbf{x}_i,y_i)\}_{i=0}^{N-1}$ is a collection of $N$ time series with their corresponding labels $y$.
    \item A Time Series Average (TSA) $\textbf{x}_{avg}=\{x_0,x_1,\ldots,x_{L-1}\}$ is a time series of length $L$ that represents the average of a part of $\mathcal{D}$.
\end{itemize}

\subsection{Time Series Similarity}

\subsubsection{Euclidean Distance (ED)}
The naive solution to define a similarity is by using the Euclidean Distance (ED). 
This metric defined in~\eqref{equ:ed} supposes that the two time series are aligned on the time axis, which is not the case most of the times.
\begin{equation}\label{equ:ed}
    ED(\textbf{x}_1,\textbf{x}_2) = \sqrt{\sum_{t=0}^{L-1}(x_{1,t} - x_{2,t})^2}.
\end{equation}
Another limitation that this similarity measure presents is that both time series should have the same length.
In case of unequal length samples in the dataset, the problem should be addressed as dicussed in~\cite{unequal-length-ts} such as padding, uniform scaling, etc.

\subsubsection{Dynamic Time Warping (DTW)}

The following measure~\cite{dtw} is a more general formulation of the ED that is: (a) independent of the time series length, and (b) aligns the two time series on the time axis.
The formulation of the DTW is presented in~\eqref{equ:dtw}.
\begin{equation}\label{equ:dtw}
    DTW(\textbf{x}_1,\textbf{x}_2) = \min_{\pi \in \mathcal{M}(\textbf{x}_1,\textbf{x}_2)} (\sum_{(i,j)\in \pi} |x_{1,i} - x_{2,j}|^{q})^{1/q},
\end{equation}
with $\mathcal{M}(\textbf{x}_1,\textbf{x}_2)$ being the set of all possible alignment paths on the time axis between $\textbf{x}_1$ and $\textbf{x}_2$.
The parameter $q$ is the order of the Minkovski distance used, if $q=2$ then the distance is set to be Euclidean.
The hypothesis in this case is that $\textbf{x}_1$ and $\textbf{x}_2$ have different lengths, $L_1$ and $L_2$, respectively.
The goal of DTW is to find the optimal path $\pi$ of length $L_{\pi}$ that minimizes the loss in~\eqref{equ:dtw}.
Some conditions should be applied on the optimal path as listed below:
\begin{itemize}
    \item $\pi_0=(0,0)$;
    \item $\pi_{L_{\pi}-1} = (L_1-1, L_2-1)$;
    \item The elements of the path should be a strictly increasing sequence in the indices $i$ and $j$ of $\pi$.
\end{itemize}

\subsubsection{Soft Dynamic Time Warping (SoftDTW)}

One issue of the DTW measure is its non-differentiability. For this reason, in~\cite{soft-dtw} the Soft Dynamic Time Warping (SoftDTW) was proposed, which is differentiable.
This differentiability exists because of the replacement of the hard $\min$ function in~\eqref{equ:dtw} by the softer version as seen in~\eqref{equ:softmin}:
\begin{equation}\label{equ:softmin}
softmin^{\gamma}(x_0,\ldots,x_{L-1}) = -\gamma . \log (\sum_{i=0}^{L-1} e^{-x_i/\gamma}).
\end{equation}
where the parameter $\gamma$ controls the smoothness of the $softmin$ function.
The smaller the value of $\gamma$, the closer the $softmin$ function is to the hard $\min$.

\subsubsection{Shape Dynamic Time Warping (ShapeDTW)}

In~\cite{shape-dtw}, a different version of DTW was proposed that, instead of aligning all the time series at the same time, aligns transformations of sub-sequences of the time series.
This is done in order to preserve the fact that the alignment between two time stamps of two different time series takes into consideration the structure of their neighborhoods.
For the mathematical definition of ShapeDTW, let us assume $\mathcal{F}$ is a descriptor function, $x_1$ and $x_2$ two univariate time series of lengths $L_1$ and $L_2$, respectively.
The first step is to extract the sub-sequences of length $l$ from $x_1$ and $x_2$ denoted by $\mathcal{X}_1$ and $\mathcal{X}_2$ represented as two multivariate time series of shape $(L_1, l)$ and $(L_2,l)$, respectively.
The second step is to extract the descriptors from the sub-sequences using $\mathcal{F}$ and produce $\mathcal{D}_1=\mathcal{F}(X_1)$ and $\mathcal{D}_2=\mathcal{F}(\mathcal{X}_2)$ of shapes $(L_1,d)$ and $(L_2,d)$, respectively, where $d$ is the target dimension.
The ShapeDTW measure comes down to the following optimization problem:
\begin{equation}\label{equ:shapedtw}
    ShapeDTW(x_1,x_2) = \min_{\pi\in\mathcal{M}(x_1,x_2)}(\sum_{(i,j)\in\pi}|\mathcal{D}_{1,i}-\mathcal{D}_{2,j}|^q)^{1/q}
\end{equation}
The above definition can simply be adapted to multivariate time series as mentioned in the original work~\cite{shape-dtw} by extracting multivariate sub-sequences and applying the descriptors on each dimension independently or by finding a suitable multivariate descriptor function.

\subsection{Time Series Averaging - Clustering}

\subsubsection{Time Series Clustering}

Given a time series dataset, usually an unlabeled one, the goal of the clustering algorithm is to learn how to group time series samples that should belong to the same class label together.
A well known clustering algorithm is the $k$-means one, which learns how to group time series samples given their distance to a cluster's centroid.
For this reason, a definition of a time series cluster centroid should be defined.

\subsubsection{Dynamic Time Warping Barycenter Averaging (DBA)}

To define an average of a collection of time series, in~\cite{dba} the usage of DTW measure was proposed in order to find the optimal average that takes into consideration the misalignment between the samples of this collection.
In other words, given two time series, the DBA algorithm defines for each time stamp its barycenter by taking the average of all the aligned values.
DBA has proven to be very effective in clustering using the $k$-means algorithm.

\subsubsection{Soft Dynamic Time Warping Barycenter Averaging (SoftDBA)}

In~\cite{soft-dtw}, authors also proposed the replacement of DTW in the DBA algorithm by using SoftDTW instead.
Our proposed approach, called SoftDBA, is shown to work better than DBA in clustering and classification.

\section{Proposed Approach}

\subsection{Shape Dynamic Time Warping Barycenter Averaging (ShapeDBA)}

ShapeDBA follows the same methodology of DBA and SoftDBA that is averaging over the aligned time stamps.
The key difference of ShapeDBA is the usage of the ShapeDTW~\cite{shape-dtw} aligning method of time series data.
The ShapeDBA algorithm can be summarized in the following steps:
\begin{itemize}
    \item \textbf{Step 1}: Initialize the average time series, for example choose a random selection of the time series set in question;
    \item \textbf{Step 2}: Find the aligned points of each time stamp of the average series with all the samples of the data. We call the time stamps of all the samples aligned with a given time stamp $t$ of the average series as $assoc_t=\{assoc_{t_0},assoc_{t_1},\ldots,assoc_{t_{A-1}}$, where $A$ is the number of associated time stamps with $t$;
    \item \textbf{Step 3}: For each time stamp $t$ of the average series, the resulting average is the $barycenter$ of $assoc_t$.
    \\Where $barycenter(assoc_{t_0}, assoc_{t_1}, \ldots, assoc_{t_{A-1}}) = \frac{1}{A}\sum_{i=0}^{A-1}assoc_{t_i}$;
    \item Repeat from Step 2 until convergence.
\end{itemize}

\subsection{Clustering with ShapeDBA}\label{sec:clustering-shapedba}

The $k$-means clustering algorithm in machine learning can be used with any time series averaging technique, coupled with any time series similarity measure.
The averaging method, i.e., ShapeDBA for instance, is used to find the centroids of each cluster during the training phase.
The similarity measure is then used to calculate the distance of each series in the data to the centroid of each cluster.

In the rest of this paper, we refer to the following coupling for applying the $k$-means clustering algorithm:
\begin{itemize}
    \item DBA: the DBA as an averaging method coupled with the DTW as a similarity measure;
    \item MED: the arithmetic mean as an averaging technique coupled with the Euclidean Distance (ED) as a similarity measure;
    MED finds iteratively the arithmetic average series, as in DBA, without taking into consideration the temporal alignment between the prototype and the samples;
    \item SoftDBA: the SoftDBA as an averaging method coupled with the SoftDTW as a similarity measure;
    \item ShapeDBA: the ShapeDBA as an averaging method coupled with the Shape\-DTW as a similarity measure.
\end{itemize}

\subsection{Implementation Efficiency}

\begin{figure}
    \centering
    \includegraphics[width=0.8\textwidth]{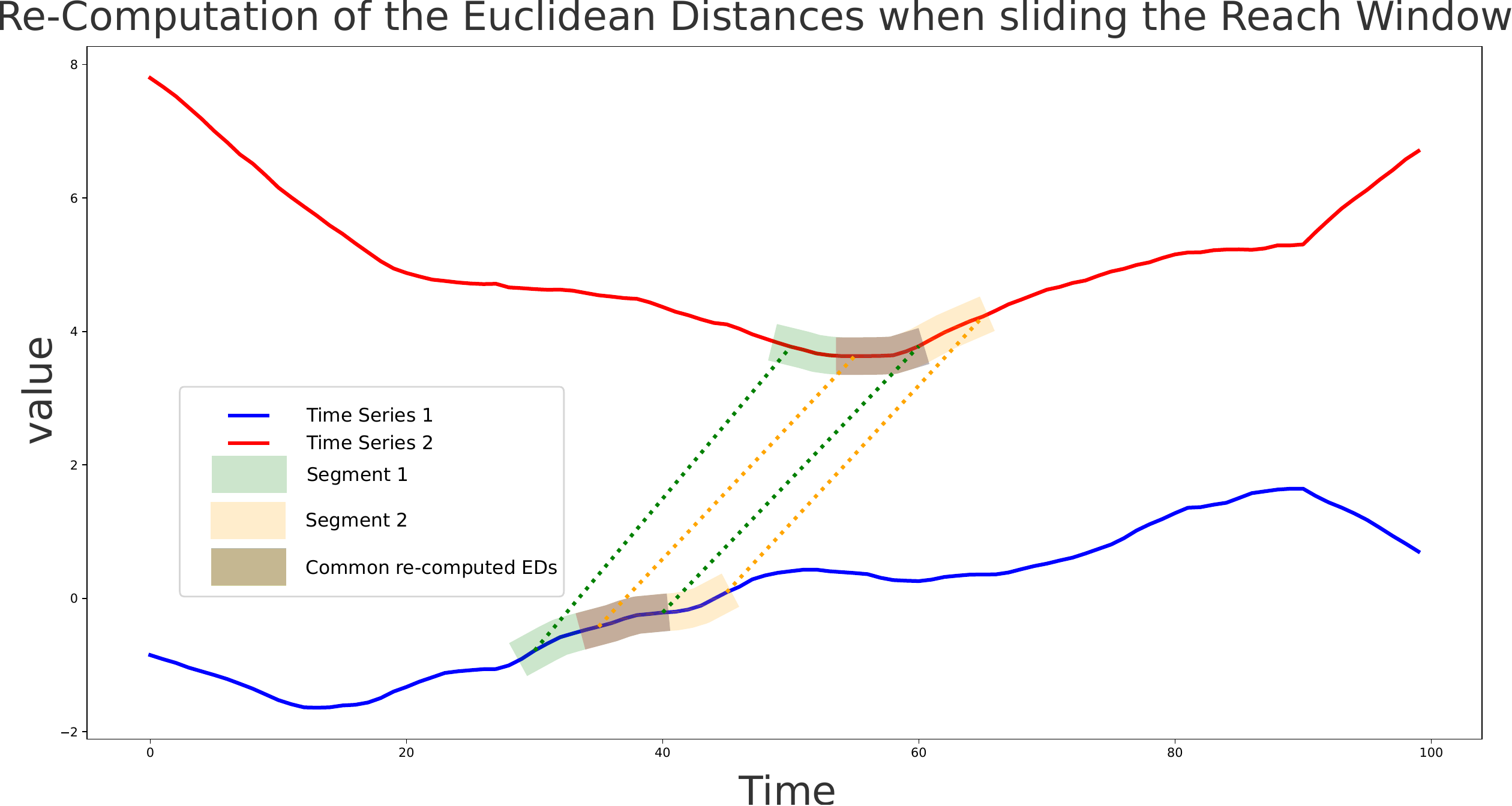}
    \caption{Computation of the ShapeDTW measure between two time series. It can be observed that the common area between the two sliding window is re-computed.}
    \label{fig:re-comuting-shapedtw}
\end{figure}

The ShapeDTW algorithm comes down to applying the original DTW similarity measure on the transformed input time series.
In the univariate case coupled with the 'identity' descriptor of each neighborhood~\cite{shape-dtw}, the transformed time series is a multivariate version.
For each time stamp, its neighborhood is added as a Euclidean vector to form a multivariate time series.
When applying the DTW similarity measure on this transformed series, the algorithm is simply computing the Euclidean distance between the channel vectors of a pair of time stamps.
This creates a computational waste when sliding the reach window as illustrated in Figure~\ref{fig:re-comuting-shapedtw}.
This problem only occurs when the descriptor is set to be the identity transformation.

\begin{figure}[t]
    \centering
    \includegraphics[width=0.8\textwidth]{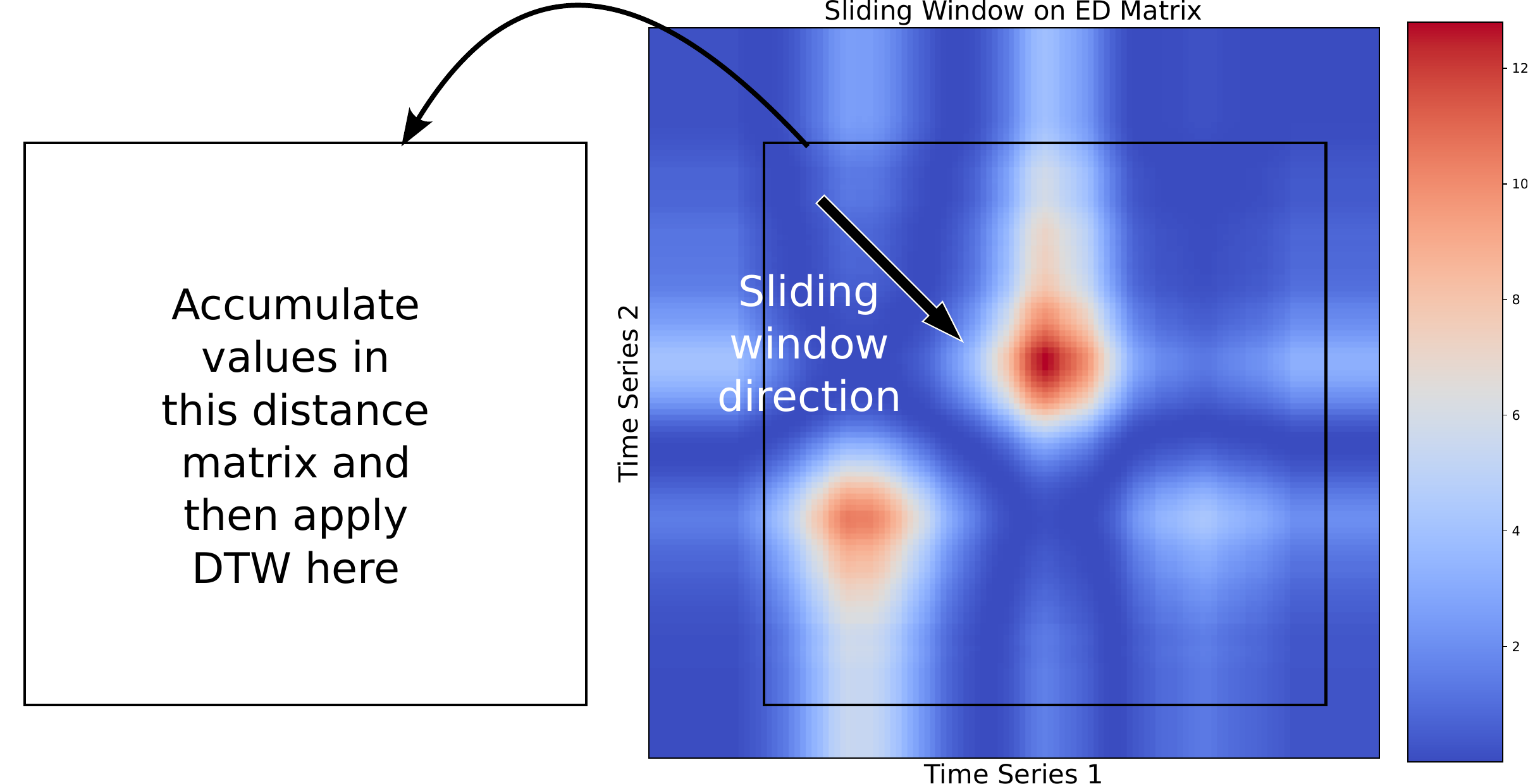}
    \caption{
    A more efficient implementation of the ShapeDTW measure when the descriptor is set to be the identity.
    Instead of applying the DTW on the multivariate transformation of the time series, a window slides on the ED matrix between the two time series.
    The captured frames are accumulated in another zero-initialized matrix on which the DTW algorithm is then applied.}
    \label{fig:eficient-shapedtw}
\end{figure}

To avoid this issue, the Euclidean pairwise distance between the two time series in question is computed as a first step.
This distance matrix is then padded with its edges values $reach / 2$ times.
We then slide a window of height and width equal to the time series lengths on this Euclidean distance matrix.
The direction of the sliding window is over the second diagonal of the distance matrix.
The results captured on the sliding window are accumulated in a zero-initialized matrix.
After accumulating all the information into the new distance matrix, we apply the DTW algorithm on the new matrix.
This implementation saves time by avoiding unnecessary computations.
A summary of this efficient implementation of the ShapeDTW can be seen in Figure~\ref{fig:eficient-shapedtw}.

\subsection{Reach Value Control}

The hyperparameter of ShapeDTW, called ``reach", controls the length of the neighborhood of each time stamp to be used for the alignment.
This value makes the ShapeDTW algorithm a general definition that includes two similarity measures: the DTW and the Euclidean distance.
For instance, on the one hand, if the reach value is set to $1$, the algorithm will behave just as the original DTW similarity measure.
This is due to the fact that the length of the neighborhood of each time stamp will be set to $1$ leading to taking into consideration only this time stamp.
On the other hand, if the reach is large enough, i.e., $\infty$, the ShapeDTW algorithm will behave just as the Euclidean distance.
This is due to the fact that for each time stamp, the neighborhood length will be larger than the time series itself.
In this work, we set the value of the reach to $30$ given it was the value used in the original paper~\cite{shape-dtw}.

\section{Results}

\subsection{Experimental Setup}

\subsubsection{Datasets}

All the experiments were conducted on 123 datasets of the UCR archive~\cite{ucr-archive}. 
The total number of datasets in the UCR archive since 2018 is 128, but five datasets were excluded from the experiments given the large length of the time series.
This was crucial given the quadratic time complexity of most of the executed algorithms with respect to the time series length.
All of the datasets were $Z$-normalized in order to have a zero mean and unit standard deviation for each time series. 
The clustering algorithms are trained on the combination of the train test splits for all the 123 datasets used in the experiments.
It is important to note that some datasets of the UCR archive are simply another train test split of an exiting dataset.
This does not occur much, which would mean that the clustering algorithm is done on the same dataset more than one time.
The source code of this work is publicly available for reproducibility~\footnote{https://github.com/MSD-IRIMAS/ShapeDBA}.

\subsubsection{Removing Bias}

A typical problem in non-deterministic estimators in machine learning is the biased performance to a given initial setup.
This problem occurs in many problems such as deep learning where the performance can be biased to an initialization of the weights.
In this clustering task, the bias in performance comes down to the initialization of the clusters before the $k$-means algorithm starts its optimization.
To avoid this bias, we do the same experiments five different times, each time with different initial clusters and present the average performance on each dataset.
However, this may raise the issue of fairness among multiple clustering algorithms experimented with.
This is due to the probable second bias of a method to a specific five initial clusters. 
To fix this bias as well, in this work the same initial clusters are used over the five experiments for all clustering algorithms. 
Given that for clustering experiments using $k$-means and $k$-shape need the initial clusters, which are usually randomly selected, it would create an issue if not all algorithms use the same initial clusters.
For this reason, we made sure that for all the experiments done, for the same dataset, all of the clustering variants used the same initial clusters.
This was done with five different initial clusters and the average performance is presented in order to remove any variance in the results.



\subsection{Qualitative Evaluation of DBA Variants}

\begin{figure}[!ht]
    \centering
    \includegraphics[width=0.8\textwidth]{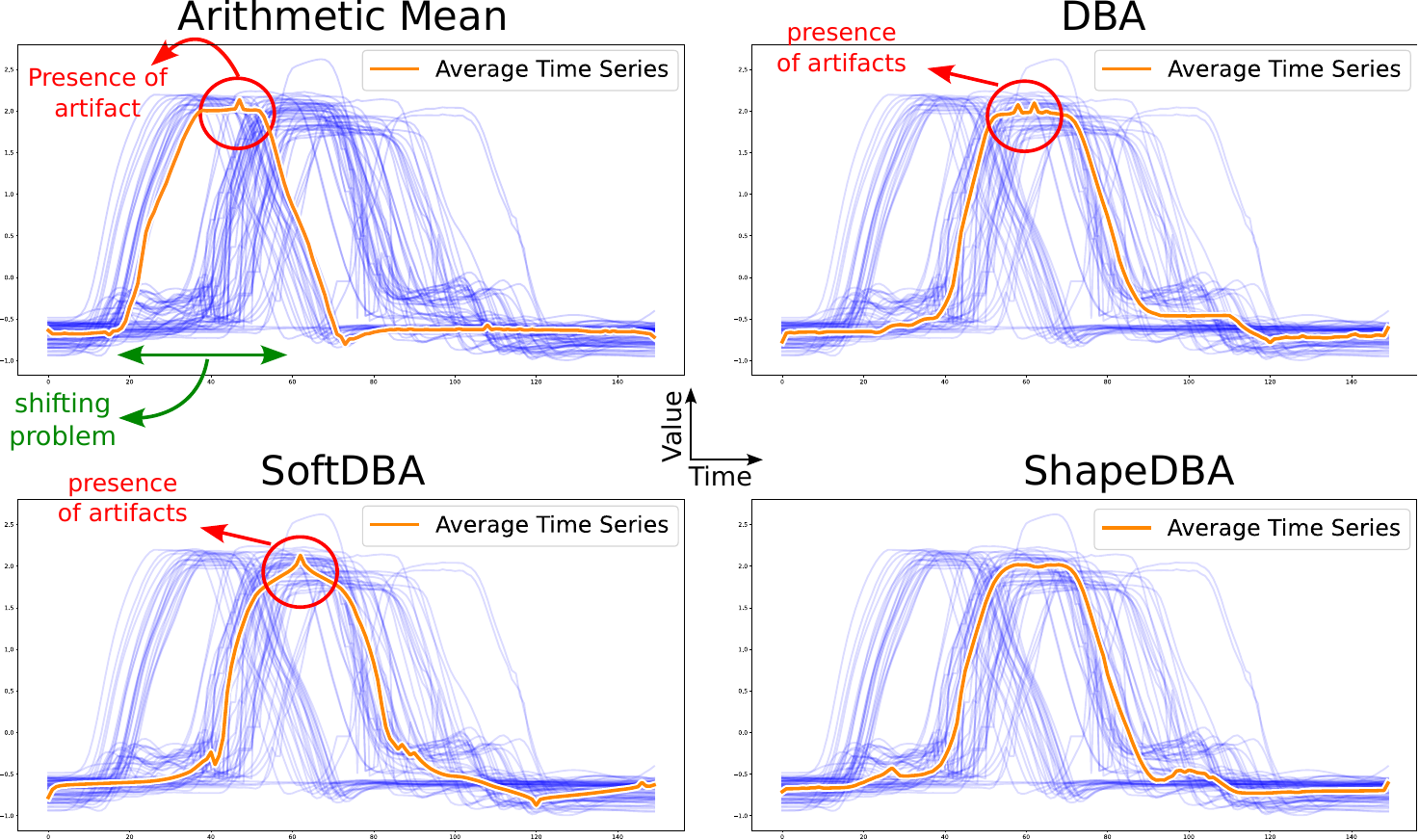}
    \caption{A qualitative evaluation of the proposed average technique compared to other approaches on a GunPoint dataset.
    The ShapeDBA algorithm is the only approach to not generate out-of-distribution artifacts.
    }
    \label{fig:averages-visulization}
\end{figure}

Given a set of time series example from the GunPointMaleVersusFemale dataset of the UCR archive, we can generate the average time series to compare and analyse the limitation of each technique.
In Figure~\ref{fig:averages-visulization}, the generated average time series is presented from a set of samples from the GunPointMaleVersusFemale dataset.
It can be seen that for the naive way of averaging, using the Euclidean distance, i.e., Arithmetic Mean, differs from all other approaches by the shifting issue.
In other words, the Arithmetic Mean does not take into consideration the time warping and miss-aligned information between the samples of the example set.

Comparing other alignment techniques with ShapeDBA, the TSA almost is placed in the same time interval.
The difference between warping methods is that DBA and SoftDBA present additional artifacts in the shape.
This results in a TSA that includes some small peaks (red circles in Figure~\ref{fig:averages-visulization}) that do not appear in the original set of time series.
ShapeDBA avoids generating this kind of artifacts given the usage of shapeDTW.
ShapeDTW's advantage is to avoid aligning a time stamp with an outlier, which is obtained thanks to the ability of the method of aligning time stamp in specific sub-sequence of the time series.
This advantage leads ShapeDBA to generate a prototype that is more likely to be randomly selected from the dataset distribution.

\subsection{Quantitative evaluation}

\subsubsection{Competitor}

In this work, we compare the proposed method to other time series averaging techniques as detailed in Section~\ref{sec:clustering-shapedba}.
The state-of-the-art model for time series clustering is $k$-shape~\cite{kshape}.
This algorithm is an improvement over the $k$-means algorithm on time series data by using a Shape Based Distance (SBD) that uses the cross-correlation between two time series instead of an alignment measure.
Until now, to the best of our knowledge, $k$-shape is the state-of-the-art and most efficient clustering method on time series data.

\subsubsection{Adjusted Rand Index (ARI)}
The Adjusted Rand Index (ARI)~\cite{ari} is a new fixed version of the original Rand Index (RI) defined in~\eqref{equ:ri}.
Given the true labels of the time series dataset $\textbf{y}$ and the predicted labels by the clustering algorithm $\hat{\textbf{y}}$, the RI is calculated as follows:
\begin{equation}\label{equ:ri}
    RI(\textbf{y},\hat{\textbf{y}}) = \dfrac{TP+TN}{TP+FP+FN+TN},
\end{equation}
where, TP and TN stand, respectively, for True Positive and True Negative, while FP and FN stand for False Positive and False Negative, respectively. 

The RI counts the number of pairs that are present in the intersection of both sets of true and predicted labels as well as the number of pairs that exist in the difference of these two sets.
This metric, however, presents a limitation: a high RI should indicate that the two clusters in question are almost identical, which is not always the case.
The RI may favor high identical clusters without taking into consideration the case where the intersection was randomly generated.
This is due to the fact that the expected value of the RI is not constant between two random clusters.
This random chance can be generated when the number of clusters becomes high enough that the probability of a pair to be in both clusters is large.
For this reason, the Adjusted Rand Index (ARI) is proposed with a scaled version that takes into account this randomness by setting the value 0.0 for the random chance.
The ARI presented in~\eqref{equ:ari} is bounded between $-0.5$ indicating no similarity and $1.0$ for a perfect similarity between the clusters.
\begin{equation}\label{equ:ari}
    ARI(\textbf{y},\hat{\textbf{y}}) = \frac{RI(\textbf{y},\hat{\textbf{y}}) - E[RI]}{1.0 - E[RI]},
\end{equation}
where $E[RI]$ is the expected value of RI.

We present in the following three different ways to compare the performance of each clustering method on the total of 123 datasets of the UCR archive.

\begin{figure}[!ht]
    \centering
    \subfloat[\centering \label{fig:1v1-med} 1V1 with Mean Euclidean Distance]{\includegraphics[width=0.4\textwidth]{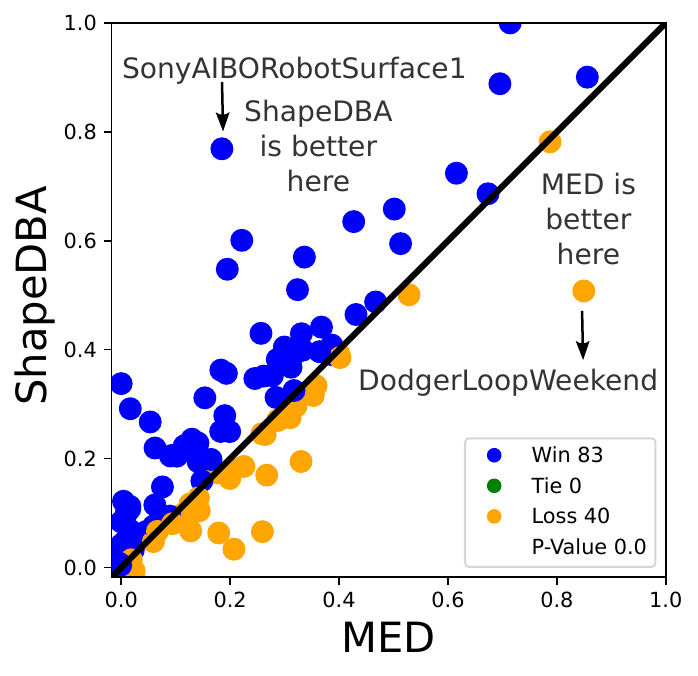}}
    \subfloat[\centering \label{fig:1v1-dbadtw} 1V1 with DBA using DTW as a metric]{\includegraphics[width=0.4\textwidth]{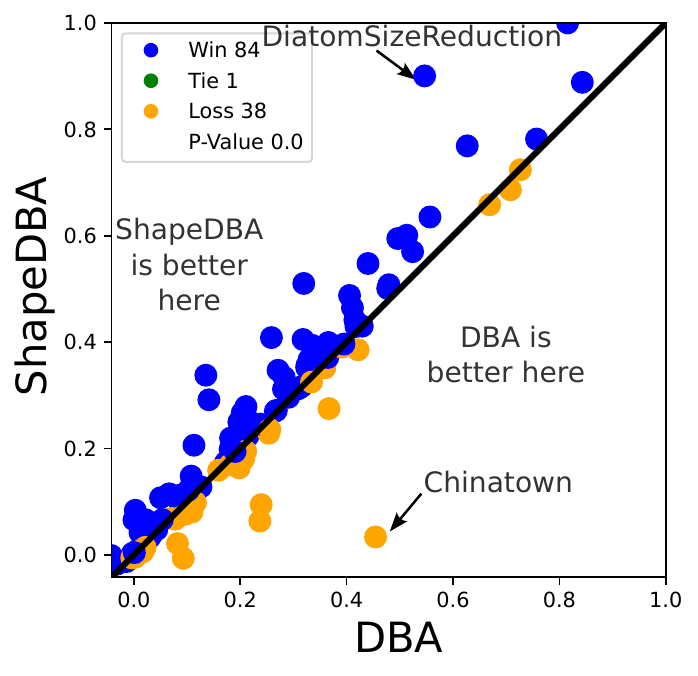}}\\
    
    \subfloat[\centering \label{fig:1v1-kshape} 1V1 with KShape]{\includegraphics[width=0.4\textwidth]{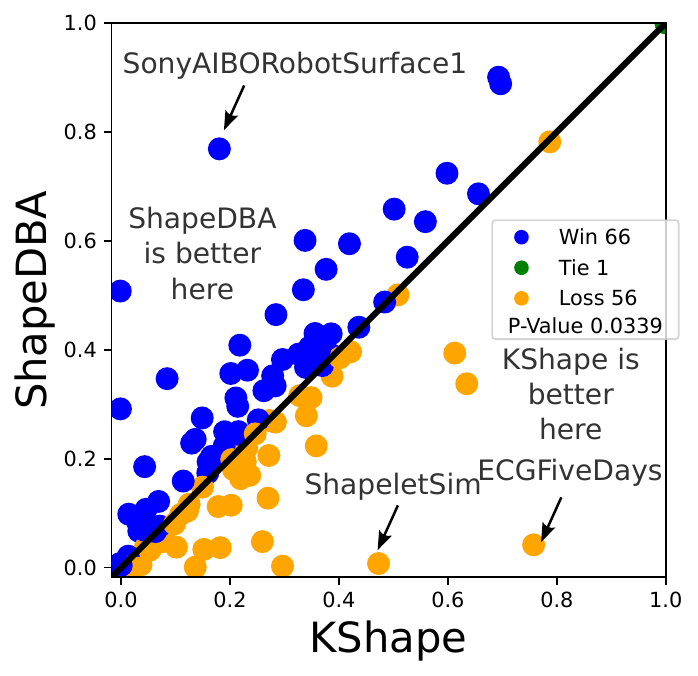}}
    \subfloat[\centering \label{fig:1v1-softdbasoftdtw} 1V1 with SoftDBA using SoftDTW as a metric]{\includegraphics[width=0.4\textwidth]{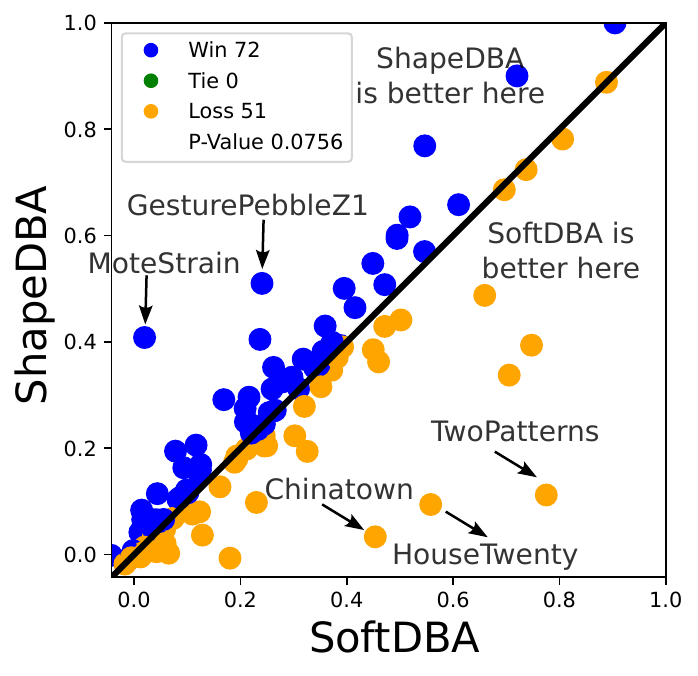}}
    \caption{1v1 Comparison between using $k$-means with ShapeDBA-ShapeDTW and other approaches from the literature using the Adjusted Rand Index clustering metric.}
    \label{fig:1v1-ari}
\end{figure}

\paragraph{One-vs-One Comparison}: In this approach, we present a scatter plot of all the pairwise comparisons between $k$-means with ShapeDBA and the approaches in the literature.
Each point visualized in Figure~\ref{fig:1v1-ari} represents one dataset, the $x$-axis presents the ARI value on this dataset using a method from the literature and the $y$-axis the ARI value using ShapeDBA.
The Win-Tie-Loss count is presented in the legend of each One-vs-One scatter plot as well as a $p$-value. This latter $p$-value is produced using the Wilcoxon Signed Rank Test~\cite{wilcoxon1992individual}.
If this $p$-value is larger than the threshold $0.05$, than the difference in performance between the comparates in question is not considered statistically significant.

It is clear from Figures~\ref{fig:1v1-med},~\ref{fig:1v1-dbadtw}, and~\ref{fig:1v1-kshape} that the usage of ShapeDBA as an averaging method in $k$-means is significantly better than the baseline, i.e., ED and DBA with $k$-means and significantly better than the state-of-the-art $k$-shape.
From Figure~\ref{fig:1v1-softdbasoftdtw} it can be seen that even though ShapeDBA presents more wins compared to SoftDBA, the difference in performance is still not significantly different.
In what follows, we show however that ShapeDBA is way faster than SoftDBA.

\paragraph{Analysing Outliers}

\begin{figure}[!ht]
    \centering
    \includegraphics[width=0.8\textwidth]{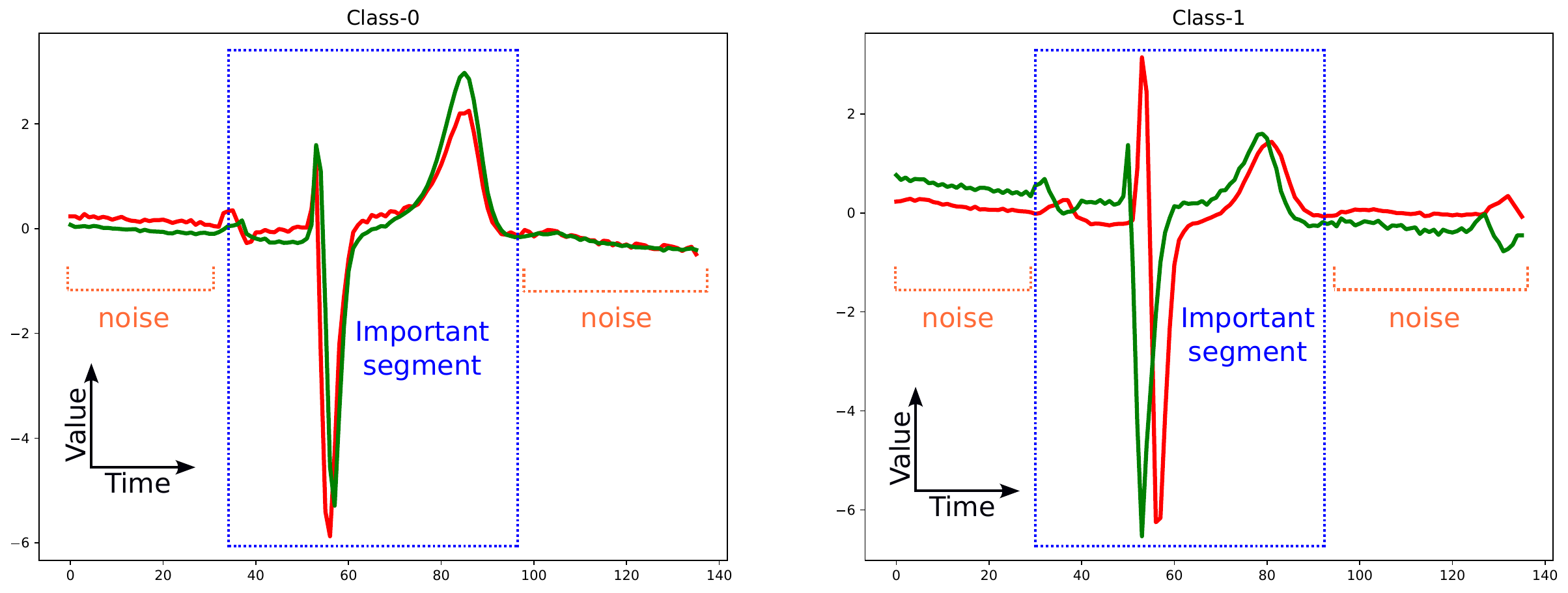}
    \caption{Two examples from each class taken from the ECGFiveDays dataset of the UCR archive. Most time stamps of this dataset represent noise and the important neighborhood of the time stamp is just in the middle of the whole time series.
    }
    \label{fig:ecg5days}
\end{figure}

Some unique outliers from the One-vs-One scatter plots are clear to favor either ShapeDBA or the other approaches.
For instance, compared to $k$-shape, ShapeDBA does not perform well (low ARI) on two datasets: ShapeletSim and ECGFiveDays.
On the one hand, given knowledge on the UCR archive datasets, we believe that no correct conclusion can be found on ShapeletSim given that this dataset is simply a simulation of random data.
On the other hand, the ECGFiveDays dataset presented in Figure~\ref{fig:ecg5days} is a unique example to show case the disadvantage of ShapeDBA.

This dataset is mostly made of noisy time stamps with an information compressed in the important segments placed in the middle of the time series as seen in Figure~\ref{fig:ecg5days}.
For this reason, ShapeDTW will be adding noise in the optimization steps. 
A clear winner on the SonyAIBORobotSurface1 dataset, however, is ShapeDBA compared to $k$-shape with almost a $0.6$ difference in the ARI.
After analysing this dataset, still no hard conclusions can be found but this is not a special case for ShapeDBA given that MED, DBA and SoftDBA perform better than $k$-shape on this dataset.
Suggesting that it is $k$-shape underperforming on this dataset.

Comparing ShapeDBA to DBA, it seems as if ShapeDBA has an advantage over the DiatomSizeReduction dataset, which suffers from the lack of training samples with only four samples per class label.

\begin{figure}[!ht]
    \centering
    \includegraphics[scale=0.5]{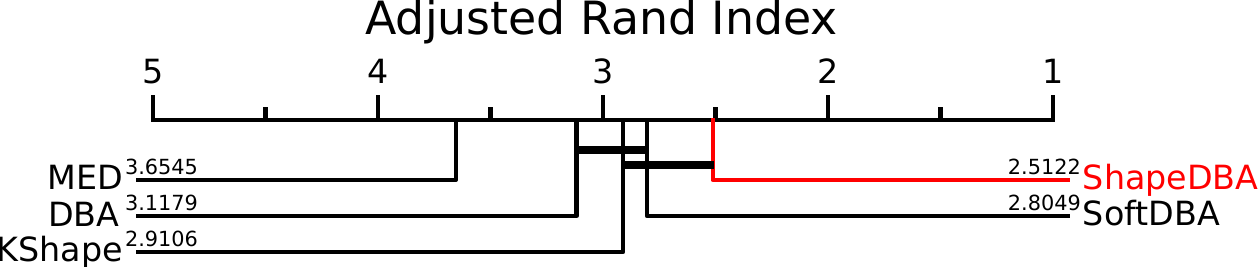}
    \caption{Critical Difference Diagram showing the average rank of the ARI score over the datasets of the UCR archive.}
    \label{fig:cdd-ari}
\end{figure}

\paragraph{Critical Difference Diagram (CDD)}:
a technique to compare multiple estimators 
by reducing the metrics on multiple datasets into a one dimensional view.
This one dimensional view is presented by using the average rank of each method on the total of the 123 datasets used.
The best performing clustering approach is the one with the lowest rank as for instance ShapeDBA in Figure~\ref{fig:cdd-ari}.
The CDD used in this work utilizes, as proposed in~\cite{benavoli2016should}, the Wilcoxon Signed-Rank Test~\cite{wilcoxon1992individual} coupled with the Holm multiple test correction~\cite{holm1979simple} in order to generate the cliques.
If a clique links a set of comparates in the CDD, this represents that the differences in performance between this set of comparates is not statistically significant.

\begin{figure}[!ht]
    \centering
    \makebox[0pt]{\includegraphics[width=\textwidth]{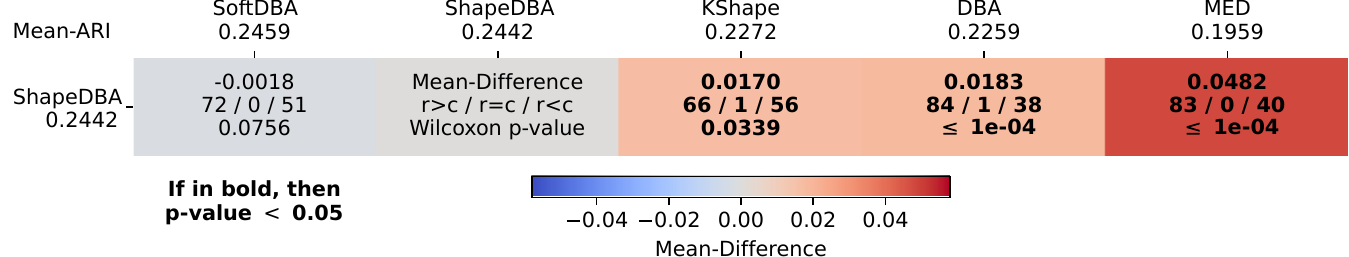}}
    \caption{A Multi-Comparison Matrix showing the proposed approach's performance compared to other approaches using a tool that is stable to the addition/removal of new classifiers.
    }
    \label{fig:mcm-ari}
\end{figure}

\begin{figure}[!ht]
    \centering
    \includegraphics[width=\textwidth]{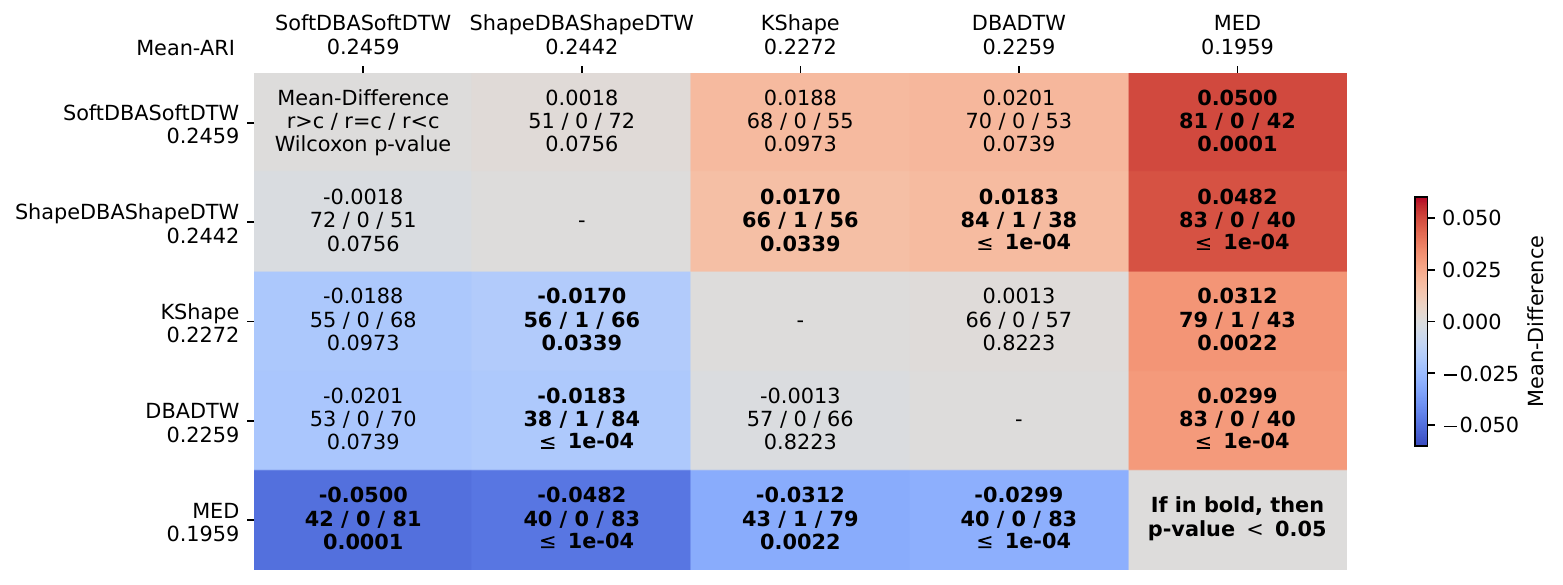}
    \caption{A Multi-Comparison Matrix showing the full One-vs-One comparison and the multi-comparates comparison between all the time series clustering approaches used and proposed in this work.}
    \label{fig:mcm-ari-all}
\end{figure}

\paragraph{Multi-Comparison Matrix (MCM)}: was proposed in~\cite{ismail2023approach} arguing that CDD has some limitations that can miss-lead the interpretation of the results.
First, one important issue with CDD as mentioned in~\cite{ismail2023approach} is the instability of the average rank. 
For instance the average rank can easily be manipulated by the addition or removal of some comparates.
For this reason, MCM proposes the usage of a descriptive statistics that does not change with this addition and removal of comparates.
This statistics is the average performance on the total of the 123 datasets used, in our case it is the average ARI over these datasets for each clustering approach.
Second, a common issue of the CDD is the usage of the multiple test correction, which is unstable to the addition and removal of comparates.
Finally, a major limitation with only using the CDD is the lack of pairwise comparison information.
The MCM proposed in~\cite{ismail2023approach} overcomes these three problems by using the average performance instead of the average rank to order the comparates, not applying a multiple test correction for the produced Wilcoxon $p$-values and presenting the pairwise comparisons between comparates.
The MCM in Figure~\ref{fig:mcm-ari} shows that SoftDBA is the winning approach given the average ARI with not much difference with the average ARI of ShapeDBA that comes in second place.
A full pairwise and multi-comparates comparison between all clustering techniques discussed in this work on the ARI metric is presented in Figure~\ref{fig:mcm-ari-all}.

In what follows, we did a computational runtime
comparison between all approaches. We show that although ShapeDBA does not outperform in significant manner SoftDBA, it is however faster.

\subsubsection{Computational Runtime}

\begin{figure}[!ht]
    \centering
    \includegraphics[scale=0.5]{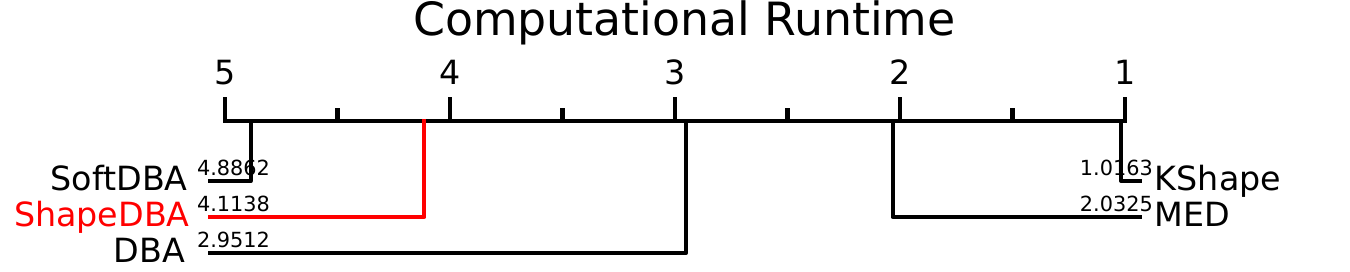}
    \caption{Critical Difference Diagram showing the average rank of the duration (in seconds) of the $k$-means algorithm over the datasets of the UCR archive.}
    \label{fig:cdd-duration}
\end{figure}

Given that all experiments were conducted on the same machine with the same environment, fairness in time computation comparison stands here.
By keeping track of the total computation time for each clustering approach, averaged over five initialization, we can apply the same comparison techniques as for the ARI.
In Figure~\ref{fig:cdd-duration}, the CDD of the computational runtime is presented.
Given that in the case of runtime, the lower the time the better, and to keep the ordering of the average rank as lower is better, we multiplied the values of the computational time by $-1$.
It is clear from the CDD plot that the fastest approach is $k$-shape and the slowest one is SoftDBA.
The reason behind the fast computation of $k$-shape is essentially because of the usage of the Fast Fourier Transform (FFT), while doing the cross-correlation between the time series.
However, with the help of the efficient implementation used in ShapeDBA, the computation is way faster than SoftDBA.

\begin{figure}[!ht]
    \centering
    \makebox[0pt]{\includegraphics[width=\textwidth]{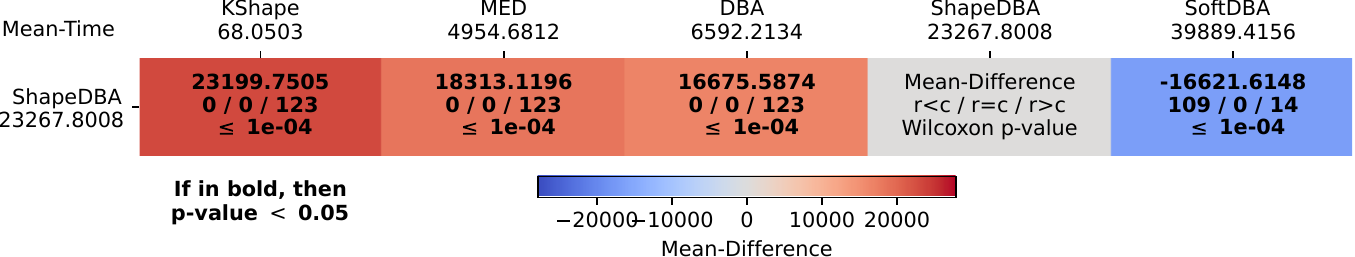}}
    \caption{A Multi-Comparison Matrix showing the proposed approach's duration (in seconds) compared to other approaches using a tool that is stable to the addition/removal of new classifiers.
    }
    \label{fig:mcm-duration}
\end{figure}

For ARI, we generated the MCM as well for the computational time comparison in Figure~\ref{fig:mcm-duration}.
On average of 123 datasets, ShapeDBA is $1.7$ times faster than SoftDBA with 109 wins for ShapeDBA in terms of computational runtime.
It is important to note that in this case of MCM, the Win-Tie-Loss count considers the lower the better.

\section{Conclusion}

In this work, we addressed the problem of Time Series Averaging (TSA) using elastic distances.
We proposed a novel TSA approach, ShapeDBA, based on the similarity measure ShapeDTW similarity measure.
We showed that ShapeDBA has the ability to preserve the shape of the true dataset distribution instead of producing spikes artifacts as other approaches.
To quantitatively evaluate the proposed approached, we provided extensive experiments on the UCR archive using the $k$-means clustering algorithm.
We show that in terms of the Adjusted Rand Index metric, our approach achieves state-of-the-art performance, while being much faster than SoftDBA that represents the current elastic state-of-the-art averaging technique.
This last observation is beneficial to help deploy time series averaging techniques in real life problems.
Finally, to avoid computation waste in our proposed ShapeDBA algorithm, we present a dynamic programming detailed implementation of the algorithm.

\section{Acknowledgements}

This work was supported by the ANR DELEGATION
project (grant ANR-21-CE23-0014) of the French
Agence Nationale de la Recherche. The authors
would like to acknowledge the High Performance
Computing Center of the University of Strasbourg
for supporting this work by providing scientific sup-
port and access to computing resources. Part of the
computing resources were funded by the Equipex
Equip@Meso project (Programme Investissements
d’Avenir) and the CPER Alsacalcul/Big Data. The
authors would also like to thank the creators and
providers of the UCR Archive.

\bibliographystyle{splncs04}
\bibliography{mybibliography}

\end{document}